\documentclass[sigconf]{acmart}

\def\BibTeX{{\rm B\kern-.05em{\sc i\kern-.025em b}\kern-.08emT\kern-.1667em\lower.7ex\hbox{E}\kern-.125emX}}
    
\copyrightyear{2024}
\acmYear{2024}
\setcopyright{acmlicensed}
\acmConference[KDD MLF '24]{KDD MLF '24: KDD Workshop on Machine Learning in Finance}{August 26, 2024}{Barcelona, Spain}
\acmBooktitle{KDD MLF '24: KDD Workshop on Machine Learning in Finance, August 26, 2024, Barcelona, Spain}
\acmPrice{}
\acmDOI{}
\acmISBN{}

\usepackage{xcolor, soul, xspace}
\newcommand{\ourdataset}{\textit{Elliptic2}\xspace}

\begin{document}

\title{The Shape of Money Laundering: Subgraph Representation Learning on the Blockchain with the Elliptic2 Dataset}

\renewcommand{\shorttitle}{The \ourdataset Dataset}

%
% The "author" command and its associated commands are used to define the authors and their affiliations.
% Of note is the shared affiliation of the first two authors, and the "authornote" and "authornotemark" commands
% used to denote shared contribution to the research.
\author{Claudio Bellei}
\authornote{These authors contributed equally to this research.}
\affiliation{%
  \institution{Elliptic}\country{}}
\email{c.bellei@elliptic.co}

\author{Muhua Xu}
\authornotemark[1]
\affiliation{%
  \institution{MIT CSAIL}\country{}}
\affiliation{%
  \institution{MIT-IBM Watson AI Lab}\country{}}
\email{muhuaxu@mit.edu}

\author{Ross Phillips}
\authornotemark[1]
\affiliation{%
  \institution{Elliptic}\country{}}
\email{r.phillips@elliptic.co}

\author{Tom Robinson}
\affiliation{%
  \institution{Elliptic}\country{}}
\email{tomrobinson@elliptic.co}

\author{Mark Weber}
\affiliation{%
  \institution{MIT Media Lab}\country{}}
\email{mrweber@mit.edu}

\author{Tim Kaler}
\affiliation{%
  \institution{MIT CSAIL}\country{}}
\affiliation{%
  \institution{MIT-IBM Watson AI Lab}\country{}}
\email{tfk@mit.edu}

\author{Charles E. Leiserson}
\affiliation{%
  \institution{MIT CSAIL}\country{}}
\affiliation{%
  \institution{MIT-IBM Watson AI Lab}\country{}}
\email{cel@mit.edu}

\author{Arvind}
\affiliation{%
  \institution{MIT CSAIL}\country{}}
\affiliation{%
  \institution{MIT-IBM Watson AI Lab}\country{}}
\email{arvind@csail.mit.edu}

\author{Jie Chen}
\affiliation{%
  \institution{IBM Research}\country{}}
\affiliation{%
  \institution{MIT-IBM Watson AI Lab}\country{}}
\email{chenjie@us.ibm.com}

%
% By default, the full list of authors will be used in the page headers. Often, this list is too long, and will overlap
% other information printed in the page headers. This command allows the author to define a more concise list
% of authors' names for this purpose.
\renewcommand{\shortauthors}{Bellei et al.}

%
% The abstract is a short summary of the work to be presented in the article.
\begin{abstract}
Subgraph representation learning is a technique for analyzing local structures (or shapes) within complex networks. Enabled by recent developments in scalable Graph Neural Networks (GNNs), this approach encodes relational information at a subgroup level (multiple connected nodes) rather than at a node level of abstraction. We posit that certain domain applications, such as anti-money laundering (AML), are inherently subgraph problems and mainstream graph techniques have been operating at a suboptimal level of abstraction. This is due in part to the scarcity of annotated datasets of real-world size and complexity, as well as the lack of software tools for managing subgraph GNN workflows at scale. To enable work in fundamental algorithms as well as domain applications in AML and beyond, we introduce \textit{Elliptic2}, a large graph dataset containing 122K labeled subgraphs of Bitcoin clusters within a background graph consisting of 49M node clusters and 196M edge transactions. The dataset provides subgraphs known to be linked to illicit activity for learning the set of ``shapes'' that money laundering exhibits in cryptocurrency and accurately classifying new criminal activity. Along with the dataset we share our graph techniques, software tooling, promising early experimental results, and new domain insights already gleaned from this approach. Taken together, we find immediate practical value in this approach and the potential for a new standard in anti-money laundering and forensic analytics in cryptocurrencies and other financial networks.
\end{abstract}

%
% The code below is generated by the tool at http://dl.acm.org/ccs.cfm.
% Please copy and paste the code instead of the example below.
%


%
% Keywords. The author(s) should pick words that accurately describe the work being
% presented. Separate the keywords with commas.

%
% A "teaser" image appears between the author and affiliation information and the body 
% of the document, and typically spans the page. 
%% \begin{teaserfigure}
%%   \includegraphics[width=\textwidth]{sampleteaser}
%%   \caption{Seattle Mariners at Spring Training, 2010.}
%%   \Description{Enjoying the baseball game from the third-base seats. Ichiro Suzuki preparing to bat.}
%%   \label{fig:teaser}
%% \end{teaserfigure}

%
% This command processes the author and affiliation and title information and builds
% the first part of the formatted document.
\maketitle

%%%%%%%%%%%%%%%%%%%%%%%%%%%%%%%%%%%%%%%%%%%%%%%%%%%%%%%%%%%%%%%%%%%%%%%%%%%%%%%%%%%%%%%%%%%%%%%%%%%%%%%%%%%%%%%%%%%%%%%%%%%%%%%%%%%%%%%
\section{Introduction}
Over the last few years, the development of graph neural networks (GNNs) has allowed the extension of deep learning methods to data that have an underlying non-Euclidian structure. GNNs offer a rich variety of models and architectures that enable the learning of meaningful representations for nodes, edges, and the entire graphs. These variants have found applications in diverse fields such as recommender systems, chemistry, traffic control, physics, and more \cite{zhou2021graph}. However, when dealing with complex graph structures, it is often possible to identify subgraphs of particular interest. The recent emergence of subgraph representation learning has allowed the prediction of subgraph properties within the larger graph structure \cite{alsentzer2020subgraph, glass}. By leveraging this approach, it is possible to gain valuable insights into the characteristics and behavior of these subgraphs within the broader background graph.

Interesting subgraphs can be found in financial graphs such as blockchain-based cryptocurrencies. Since their inception with Bitcoin \cite{Nakamoto}, one of the primary features of cryptocurrency is the immutable and transparent record of all transactions ever logged on the network, while maintaining pseudonymity for users. Combining this public information with knowledge about the presence of licit and illicit services on the network, cryptocurrency intelligence companies have emerged to provide Anti-Money-Laundering (AML) solutions tailored to the cryptocurrency domain. Whereas the pseudonymity of Bitcoin is an advantage for criminals, the public availability of data is a key advantage for those within law enforcement agencies and financial institutions looking to identify and investigate financial crime.

The dataset here presented identifies subgraphs of interest within the AML lens of a cryptocurrency intelligence company: some subgraphs carry anomalous signatures of money laundering activity, while others (the vast majority) appear to be transferring flows of bitcoins among licit services. The challenge is being able to make predictions on subgraphs that have not yet been labeled, and potentially will never be labeled using standard methodologies, to allow AML solutions to make predictions even in the presence of unknowns by making exclusive use of public blockchain data.

%%%%%%%%%%%%%%%%%%%%%%%%%%%%%%%%%%%%%%%%%%%%%%%%%%%%%%%%%%%%%%%%%%%%%%%%%%%%%%%%%%%%%%%%%%%%%%%%%%%%%%%%%%%%%%%%%%%%%%%%%%%%%%%%%%%%%%%
\section{A massive dataset for subgraphs}
\label{dataset}
Available real-world datasets for subgraph representation learning include \textit{PPI-BP}, \textit{HPO-METAB}, \textit{HPO-NEURO}, and \textit{EM-USER}. Detailed properties for each can be found in \cite[Appendix B]{alsentzer2020subgraph}. The largest background graphs among these datasets constitute 100K nodes and 5M edges, with the number of subgraphs varying between 324 and 4,000 and with nodes per subgraph ranging between 10 and 155. To our knowledge, these are the largest subgraph datasets available, but they are not so large indeed. This limits scientific advancement in this field because resolving scalability constraints and other challenges common to massive network structures are indeed central to the promise of GNNs in many practical settings.

To enable research on subgraph learning on a much larger scale, we present \ourdataset, a fully connected network of Bitcoin addresses and transactions between them comprising\textbf{ a background graph almost three orders of magnitude larger than any available real-world dataset that we know of}. Within this background graph are many small subgraphs labeled as suspicious or licit, the task at hand being the binary classification of suspicious subgraphs. The release of this dataset follows our publication of a standard graph dataset in 2019, which contained Bitcoin transactions and focused on node classification. Both datasets enable the advancement of research in scalable Graph Neural Networks and their applications for anti-money laundering in cryptocurrency.

%%%%%%%%%%%%%%%%%%%%%%%%%%%%%%%%%%%%%%%%%%%%%%%%%%%%%%%%%%%%%%%%%%%%%%%%%%%%%%%%%%%%%%%%%%%%%%%%%%%%%%%%%%%%%%%%%%%%%%%%%%%%%%%%%%%%%%%
\subsection{Retrospect on \textit{Elliptic1}}
In 2019, we published a labeled graph dataset of Bitcoin transactions named \textit{The Elliptic Data Set} (hereafter referred to as \textit{Elliptic1}) on Kaggle \cite{elliptickaggle2019} with an accompanying paper \cite{weber2019} demonstrating through experimental results on how GNNs can be used to extract hidden relational information that can be fed to classification models for significant performance boosts.
%
% To anonymize identity
% In 2019, a labeled graph dataset of Bitcoin transactions named \textit{The Elliptic Data Set} (hereafter referred to as \textit{Elliptic1}) was published on Kaggle \cite{elliptickaggle2019} with an accompanying paper \cite{weber2019} demonstrating through experimental results on how GNNs can be used to extract hidden relational information that can be fed to classification models for significant performance boosts.
The dataset comprised 204K node transactions with 166 features and 234K directed edge payment flows. Approximately 2\% of the node transactions were labeled as illicit and 21\% labeled as licit. The task presented was a binary node classification task, predicting whether a node transaction was broadcasted to the Bitcoin network by a licit or illicit entity. As of this publication, the dataset has been viewed more than 100K times and downloaded almost 10K times, with the paper receiving approximately 400 citations. Traction in both the machine learning and AML communities motivated us to publish an even larger dataset we call \ourdataset with a modified structure and the addition of subgraph labels to enable subgraph classification as a potentially powerful new tool for AML professionals. While our own novel methods on this new dataset are forthcoming, we offer this dataset to the community now in the interest of scientific advancement and the public good.

%moved here for better look
\begin{table*}[t]
  %\resizebox{\linewidth}{!}{%
  \label{tab:background}
  \begin{tabular}{c|c|c|c|c|c}
    \small{\# Nodes} & \small{\# Edges} & \small{\# Subgraphs} & \small{\# Node features} & \small{\# Edge features} & \small{\# Classes}\\ 
    \midrule
    49,299,864 & 196,215,606 & 121,810 & 43 & 95 & 2\\
    \bottomrule
  \end{tabular}
  %}
  \vspace{1em}

  %\resizebox{\linewidth}{!}{%
  \begin{tabular}{c|c|c|c|c|c}
    % \toprule
    \small{Class} & \small{\# Subgraphs} & \small{Average \# Nodes} & \small{Median \# Nodes} & \small{Min \# Nodes} & \small{Max \# Nodes} \\ 
    \midrule
    \small{Licit} & 119,047 & 3.65 & 3 & 2 & 296 \\
    \hline
    \small{Suspicious} & 2,763 & 3.79 & 3 & 2 & 30 \\    
    \bottomrule
  \end{tabular}
  %}
  \vspace{1em}
  
  \caption{Dataset properties for the background graph (top) and for the licit and suspicious subgraphs (bottom).}
  \label{tab:dataset}

\end{table*}

%%%%%%%%%%%%%%%%%%%%%%%%%%%%%%%%%%%%%%%%%%%%%%%%%%%%%%%%%%%%%%%%%%%%%%%%%%%%%%%%%%%%%%%%%%%%%%%%%%%%%%%%%%%%%%%%%%%%%%%%%%%%%%%%%%%%%%%
\subsection{Introducing \ourdataset}
Our task with \ourdataset is to combat financial crime by identifying whether a specific flow of bitcoins may be linked to money laundering activity, specifically the attempt to convert the profits gained from illegal actions into fiat currency or other cryptocurrencies through a legitimate service. For the discussion that follows, some useful definitions are needed.

\begin{definition}
A \textit{cluster} is a set of Bitcoin addresses thought to be controlled by a single person or organization. 

A \textit{licit} cluster is owned by a ``licit'' entity (exchange, wallet provider, miner, licit service, etc.). An \textit{illicit} cluster is owned by an ``illicit'' entity (dark market, scam, hack, etc.). Clusters that are neither licit nor illicit are deemed \textit{unknown} (e.g., unlabeled clusters).

A \textit{suspicious} path is a path that connects an illicit cluster to a licit cluster. An \textit{illicit} path is a path that connects an illicit cluster to an illicit cluster. A path that is neither licit nor suspicious nor illicit is considered \textit{neutral}.
\end{definition}

%%%%%%%%%%%%%%%%%%%%%%%%%%%%%%%%%%%%%%%%%%%%%%%%%%%%%%%%%%%%%%%%%%%%%%%%%%%%%%%%%%%%%%%%%%%%%%%%%%%%%%%%%%%%%%%%%%%%%%%%%%%%%%%%%%%%%%%
\subsubsection{Assumptions}
The definitions above rest on the following assumption: a path on the blockchain connecting an illicit cluster to a licit cluster without a change of ownership of the funds likely represents the activity of money laundering by a criminal person or organization. The idea is that criminals who intend to deposit their funds at a legitimate service will try to evade detection on the blockchain, creating a distinct subgraph with associated ``shape'' and features that a machine learning model should be able to identify.

%%%%%%%%%%%%%%%%%%%%%%%%%%%%%%%%%%%%%%%%%%%%%%%%%%%%%%%%%%%%%%%%%%%%%%%%%%%%%%%%%%%%%%%%%%%%%%%%%%%%%%%%%%%%%%%%%%%%%%%%%%%%%%%%%%%%%%%
\subsubsection{Graph construction}
\begin{figure}[h]
  \centering
  \includegraphics[width=\linewidth]{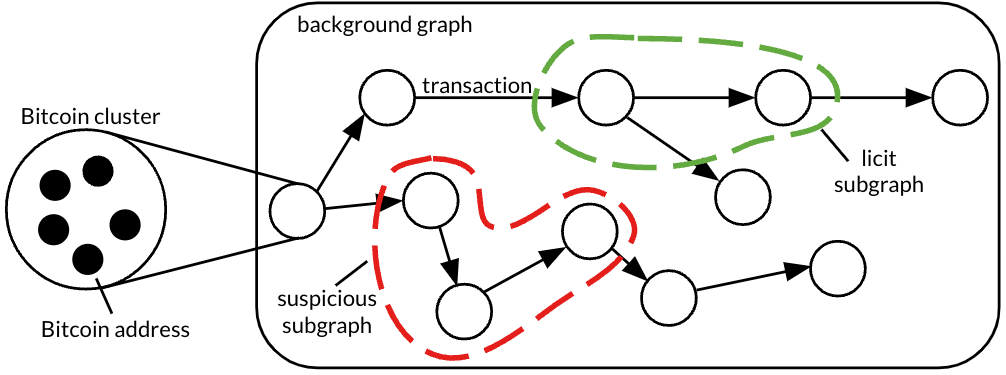}
  \caption{Illustrative example of the dataset, with its background graph and annotated subgraphs. Each node represents a Bitcoin cluster, a collection of Bitcoin addresses controlled by the same entity, and each edge representing a transaction between them.}
  \label{fig:clusters}
\end{figure}
Figure \ref{fig:clusters} shows the components of the dataset, with each node in the graph representing a Bitcoin cluster, and edges representing transactions between them. Clustering addresses is done applying well known \cite{androulaki2013,ron2013,reid2013,harrigan2016,meiklejohn2016,moser2022} as well as proprietary heuristics, human analysis and domain expertise. While annotating the subgraphs, for intellectual property reasons only a subset of Elliptic's cluster labels were used.  

The dataset is built following the steps provided below.
\begin{enumerate}
  \item Define the time window, the maximum number of hops to be traversed, and an early stopping condition when a change of ownership is likely to happen (e.g., when an unknown cluster with high activity is found during traversal).
  \item Within the time window, determine the largest connected component (background graph).
  \item For each labeled node in the background graph, obtain its outgoing transactions, providing a starting point for graph traversal.
  \item Traverse the graph until one of the following conditions is true: I. a labeled node found; II. the length traversed is larger than the maximum number of hops allowed; III. the early stopping condition is satisfied.
  \item For each path found at the previous step, define it as ``suspicious'' if illicit  $\rightarrow$  licit, ``licit'' if licit  $\rightarrow$  licit, ``illicit''  if illicit $\rightarrow$  illicit, or ``neutral'' if illicit/licit $\rightarrow$  unknown.
  \item Using the above paths, run connected components to keep only subgraphs that can be labeled, i.e., those that only have logically consistent paths (licit + neutral or suspicious + illicit + neutral).
  \item Annotate subgraphs for the final output, keeping only \textit{unknown} nodes that are part of either licit (licit subgraphs) or suspicious (suspicious subgraphs) paths.
\end{enumerate}
The time window was chosen to be 1 year of blockchain data, and the maximum number of hops was 6 for computational reasons. The transactions at step (3) were capped to a maximum for each cluster, to help balance the dataset for different actors with varying blockchain activity.

\begin{figure}[h]
  \centering
  \includegraphics[width=\linewidth]{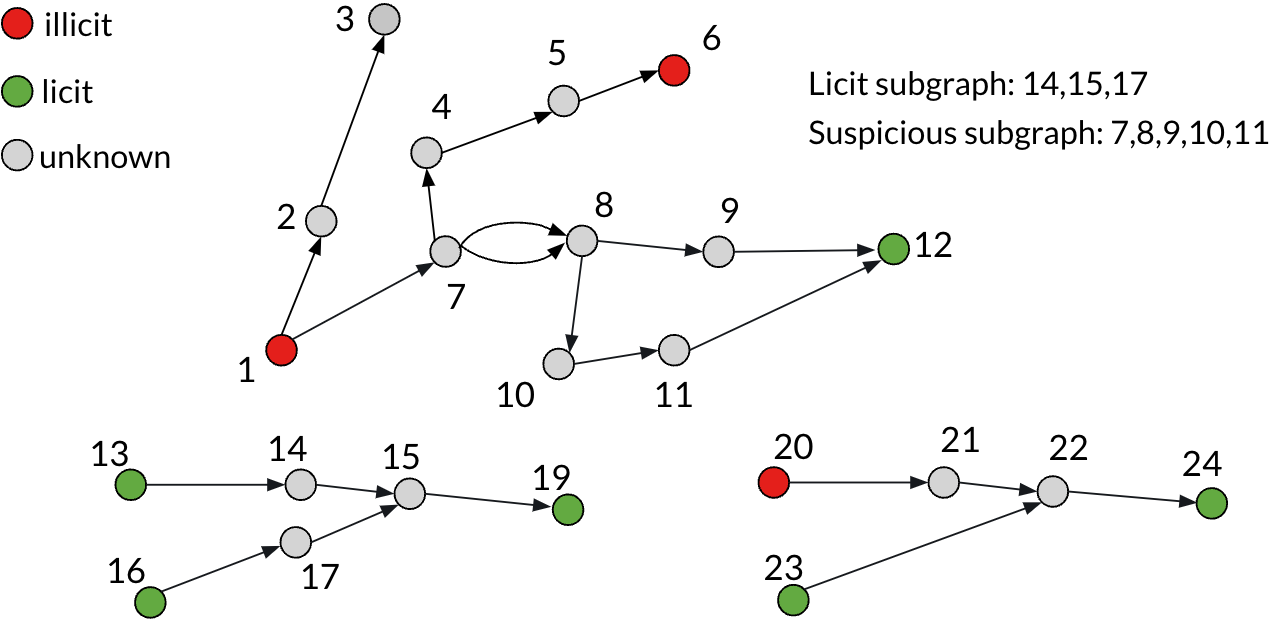}
  \caption{Construction of the dataset requires labeling paths first and then labeling subgraphs. In the example above, there are 3 licit paths (I. 13 $\rightarrow$ 14 $\rightarrow$ 15 $\rightarrow$  19; II. 16 $\rightarrow$ 17 $\rightarrow$ 15 $\rightarrow$ 19; III. 23 $\rightarrow$ 22 $\rightarrow$ 24), 1 illicit path (1 $\rightarrow$ 7 $\rightarrow$ 4 $\rightarrow$ 5 $\rightarrow$ 6), 3 suspicious paths (I. 1 $\rightarrow$ 7 $\rightarrow$ 8 $\rightarrow$ 9 $\rightarrow$ 12; II. 1 $\rightarrow$ 7 $\rightarrow$ 8 $\rightarrow$ 10 $\rightarrow$ 11 $\rightarrow$ 12; III. 20 $\rightarrow$ 21 $\rightarrow$ 22 $\rightarrow$ 24), and 1 neutral path (1 $\rightarrow$ 2 $\rightarrow$ 3). The result is one licit subgraph and one suspicious subgraph (note that the subgraph 21,22 is unlabeled as it is made of both a suspicious and a licit path).}
  \label{fig:labelling}
\end{figure}

% figure is moved here for better look
\begin{figure}[h]
  \centering
  \includegraphics[width=0.7\linewidth]{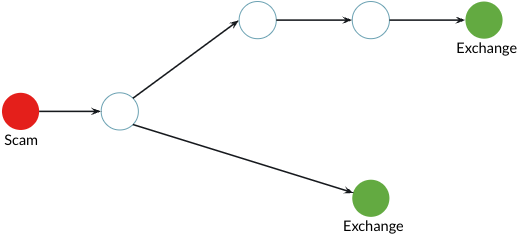}
  \caption{Example of a size-3 suspicious subgraph, connecting the profits of a scam to two exchanges (Note: the label categories are not available in the dataset).}
  \label{fig:scam}
\end{figure}

%%%%%%%%%%%%%%%%%%%%%%%%%%%%%%%%%%%%%%%%%%%%%%%%%%%%%%%%%%%%%%%%%%%%%%%%%%%%%%%%%%%%%%%%%%%%%%%%%%%%%%%%%%%%%%%%%%%%%%%%%%%%%%%%%%%%%%%
\subsubsection{Overview of \ourdataset}
An overview of some of the properties of the dataset can be found in Table \ref{tab:dataset}. The dataset is made of a background graph of 49M clusters and 196M edges. A subset of the background graph consists of labeled (licit/suspicious) subgraphs, as shown in Figure \ref{fig:clusters}. Note that the only labels present in the dataset are those associated with these subgraphs, and that the vast majority of the nodes in the graph do not belong to labeled subgraphs. Only about 2\% of the subgraphs are labeled ``suspicious'' with the remainder being labeled ``licit'', thus introducing the challenge of a significant class imbalance. On average, the size of the subgraphs is small but the tail of the distribution is significant. An example of a size-3 suspicious subgraph is presented in Figure \ref{fig:scam}. 

Each node in the background graph has 43 features, and each edge 95 features. The node features include, among others, the node size (number of addresses) and number of transactions. Both are examples of features that can vary by several orders of magnitude (from 1 to millions). The edge features include the transaction volume, fee, timestamp. To protect the intellectual property of Elliptic, most features were categorized into bins. The number of bins used varied between features. This process converted continuous numerical features into ordinal ones.

%%%%%%%%%%%%%%%%%%%%%%%%%%%%%%%%%%%%%%%%%%%%%%%%%%%%%%%%%%%%%%%%%%%%%%%%%%%%%%%%%%%%%%%%%%%%%%%%%%%%%%%%%%%%%%%%%%%%%%%%%%%%%%%%%%%%%%%
\section{Usage of \ourdataset}

\subsection{The anti-money laundering perspective}
The setup of \ourdataset is tailored to the regulatory compliance constraints that companies interacting with cryptocurrency must adhere to. As in the traditional AML setting, companies receiving funds are usually required to make a determination regarding the legitimacy of said funds with regard to potential connections to suspicious activities or organizations. For this purpose, the task proposed with \ourdataset is a binary classification with the objective of determining whether a subgraph is suspicious or not. The transparency of Bitcoin is helpful in that it allows for open, privacy-preserving forensic analysis of observable patterns (such as subgraph shapes) on the public blockchain. Utilizing advanced machine learning techniques to reduce label dependence, the promise of accurate anti-money laundering in cryptocurrency at scale seems directionally attainable. Although some ``typologies'' of money laundering in cryptocurrency have been identified that can help tackle this problem \cite{elliptictypologies2023}, a general solution to it is still difficult to achieve, for example in ``low profile'' cases where timeliness is important (such as when scammers try to quickly cash out the proceeds of their crime).

%%%%%%%%%%%%%%%%%%%%%%%%%%%%%%%%%%%%%%%%%%%%%%%%%%%%%%%%%%%%%%%%%%%%%%%%%%%%%%%%%%%%%%%%%%%%%%%%%%%%%%%%%%%%%%%%%%%%%%%%%%%%%%%%%%%%%%%
\subsection{The AI perspective}
The \ourdataset dataset is a timely contribution to the AI community, as graph representation learning research gained explosive traction over the years and demands real-life, large-scale, and challenging datasets to benchmark methods and models, typically GNNs and emergingly graph transformers~\cite{Dwivedi2021}. Example benchmark contributions include Benchmarking GNNs~\cite{Dwivedi2020}, which offers an extensible framework for reproducible benchmarking; and the Long Range Graph Benchmark~\cite{Dwivedi2022}, which designs graph datasets that can test a model's capability in reasoning over long-range interactions of nodes. These contributions generally do not involve massive graphs, for which the Open Graph Benchmark~\cite{Hu2020} is an alternative example, composed of several largest graphs widely used today. These graphs include ogbn-papers100M and MAG240M~\cite{Hu2021}, which support not only the development of machine learning models but also the research on system design. 

\ourdataset is a similarly large-scale dataset that can be used to benchmark the scalability of graph models and the efficiency of training systems. It represents a real-life use case in the financial domain---cryptocurrency---for which there are rarely public datasets available for benchmarking.

Besides the large scale and the unique domain, more importantly, \ourdataset supports the research of an emerging topic: subgraph representation learning. Traditionally, graph benchmark datasets are used to perform node-level tasks (classification and regression), graph-level tasks (similar), and edge-level tasks (link prediction). Instead, \ourdataset can be used to perform subgraph-level tasks. Subgraph classification is mathematically defined in the following.

\begin{definition}
  Given a graph $G$, let $S^G$ denote a subgraph of it and use $\subset$ to denote the subgraph relation; i.e., $S^G \subset G$. Let $\texttt{Label}$ be the label space, $\texttt{Train}$ be the set of training indices, and $\texttt{Test}$ be the set of test indices. Then, the problem of \emph{subgraph classification} is that given a collection of labeled training subgraphs $\{(S^G_i, y_i) \mid S^G_i \subset G, \, y_i \in \texttt{Label}, \, i \in \texttt{Train}\}$, predict the labels of the subgraphs from the test set $\{ S^G_j \mid S^G_j \subset G, \, j \in \texttt{Test}\}$. Note that each subgraph may be disconnected and different subgraphs may overlap.
\end{definition}

Subgraph classification is relatively less studied, compared with node/graph classification. A straightforward approach is to apply a graph model (e.g., GNN), obtain node representations, perform pooling (e.g., averaging) over the subgraph nodes, and apply a prediction head to predict the subgraph label. While being a common baseline, this approach is outperformed by several recently proposed methods~\cite{Adhikari2018,alsentzer2020subgraph,glass}. \ourdataset is a useful testbed for new methods regarding performance and scalability benchmarking.

%%%%%%%%%%%%%%%%%%%%%%%%%%%%%%%%%%%%%%%%%%%%%%%%%%%%%%%%%%%%%%%%%%%%%%%%%%%%%%%%%%%%%%%%%%%%%%%%%%%%%%%%%%%%%%%%%%%%%%%%%%%%%%%%%%%%%%%
\section{Results}
We experimented with three subgraph classification methods on \ourdataset: a conventional GNN trained with the subgraphs as independent graphs (named GNN-Seg), Sub2Vec \cite{Adhikari2018}, and GLASS \cite{glass}. Sub2Vec uses random walk samples within the subgraphs to build subgraph embeddings. Both GNN-Seg and Sub2Vec neglect the background graph, which may provide important information complementing the subgraphs themselves (e.g., the edges crossing a subgraph and the background graph), while GLASS makes use of it.

The GNN architecture of GLASS employs basic message-passing layers, supplemented with additional linear layers for node labeling, and is configured to have two layers. Hyperparameters mostly follow the default configuration provided by the authors of GLASS, with a slight tuning of the batch size (4000) and the learning rate (0.001) to improve training speed and quality.

We performed a random 80:10:10 split for training, validation, and testing, respectively. The experiments were conducted on a Linux server with 160 CPU cores and 1.2TB of RAM. We did not use GPUs, because the dataset was too large for the GPU memory to host all graph and intermediate data. For the same reason, we did not use the node/edge features. See Section~\ref{sec:scale} for a discussion on how one can perform training and inference for datasets of such a large size, by incorporating neighborhood sampling and using a scalable training sytem.

\begin{table}[h]
\begin{tabular}{c|ccc|ccc}
& \multicolumn{3}{c|}{Train}& \multicolumn{3}{c}{Test}  \\
Method & \multicolumn{1}{c|}{f1} & \multicolumn{1}{c|}{pr\_auc} & roc\_auc               & \multicolumn{1}{c|}{f1} & \multicolumn{1}{c|}{pr\_auc} & roc\_auc   
\\ \midrule
{GNN-Seg} & \multicolumn{1}{c|}{0.397}  & \multicolumn{1}{c|}{0.024}    & {0.538} & \multicolumn{1}{c|}{0.398}   & \multicolumn{1}{c|}{0.026}   & { 0.537} \\ \midrule
{Sub2Vec} & \multicolumn{1}{c|}{0.964} & \multicolumn{1}{c|}{0.145} & {0.706} & \multicolumn{1}{c|}{0.944} & \multicolumn{1}{c|}{0.022}   & {0.496} \\ \midrule
GLASS  & \multicolumn{1}{c|}{0.937} & \multicolumn{1}{c|}{0.224} & 0.915         & \multicolumn{1}{c|}{0.933} & \multicolumn{1}{c|}{0.208}   & 0.889              \\\bottomrule
\end{tabular}
\vspace{1em}
\caption{Performance of different subgraph classification methods on \ourdataset.}
\label{tab:result_auc}
\end{table}

Table \ref{tab:result_auc} compares the performance of the three methods under three evaluation metrics: micro-averaged F1, PR-AUC, and ROC-AUC. We see that GLASS significantly outperforms the other two approaches under the PR-AUC and ROC-AUC metrics, while being competitive on F1. Because of the high imbalance of the labels, the AUC metrics reflect a more fluid choice of the labeling threshold; thus, a higher performance on these metrics translates to a stronger benefit when coroborating the ranked predictions by hand. We see that GLASS is able to reliably predict suspicious subgraphs, yielding consistent performance on both training and test sets. This notable gain demonstrates not only the feasibility of obtaining insights from the dataset but also the importance of using the background graph for model training.

\begin{table}[h]
\begin{tabular}{c|c|c}
       & {  PP}  & {  PN}     \\ \midrule
{  AP} & {  245} & {  46}     \\ \midrule
{  AN} & {  765} & {  11125}  \\ \bottomrule
\end{tabular}
\vspace{1em}
\caption{Confusion matrix obtained by the GLASS method. Positive is the suspicious class.}
\label{tab:result_conf}
\end{table}

To provide a more comprehensive picture of the predictability of GLASS in the context of class imbalance, we show the confusion matrix in Table \ref{tab:result_conf}. At least one in four predicted suspicious subgraphs is truly suspicious, and 85\% of the suspicious subgraphs are correctly detected, indicating that GLASS maintains reasonable false positive and false negative rates.

In terms of computation, Sub2Vec has a lengthy pre-processing stage that takes around seven hours to finish but requires very little time for model training, whereas GNN-Seg and GLASS require heavy computation in the training of the GNN model. Overall, all methods were able to converge in several days of training with an inference time that was less than 8 hours.

%%%%%%%%%%%%%%%%%%%%%%%%%%%%%%%%%%%%%%%%%%%%%%%%%%%%%%%%%%%%%%%%%%%%%%%%%%%%%%%%%%%%%%%%%%%%%%%%%%%%%%%%%%%%%%%%%%%%%%%%%%%%%%%%%%%%%%%
\section{Corroborating the trained model}
Based on the results in the preceding section, we predicted suspiciousness of additional, unlabeled subgraphs by using the trained GLASS model.

%%%%%%%%%%%%%%%%%%%%%%%%%%%%%%%%%%%%%%%%%%%%%%%%%%%%%%%%%%%%%%%%%%%%%%%%%%%%%%%%%%%%%%%%%%%%%%%%%%%%%%%%%%%%%%%%%%%%%%%%%%%%%%%%%%%%%%%
\subsection{Insights from a cryptocurrency exchange service}
A cryptocurrency exchange was asked to validate the model predictions using their own off-chain insights. Specifically, 52 subgraphs deemed suspicious and which ended with deposit transactions to the exchange were chosen, since the exchange would hold information on the account holders that received these deposits. These deposit transactions were shared with the exchange. On review, the exchange found that 14 of the 52 accounts receiving these deposits were potentially involved in illicit activity, based on their own off-chain insights, obtained through customer due diligence and other off-chain information. Notably, of these 14 accounts, 8 had been definitively associated with ‘money laundering’ or ‘fraud’. Despite the exchange having no positive indications that the remaining 38 accounts were involved in illicit activity, this does not exclude the possibility that they were.

According to the exchange, less than 0.1\% of customer accounts have been associated with money laundering or fraud, whereas at least 26.9\% of the accounts highlighted by the model predictions were found to have such an association. This suggests that the empirical precision of our model significantly surpasses that of a naïve random model.

%%%%%%%%%%%%%%%%%%%%%%%%%%%%%%%%%%%%%%%%%%%%%%%%%%%%%%%%%%%%%%%%%%%%%%%%%%%%%%%%%%%%%%%%%%%%%%%%%%%%%%%%%%%%%%%%%%%%%%%%%%%%%%%%%%%%%%%
\subsection{Identifying the source of funds in suspicious subgraphs}
Another approach to verifying the model predictions is to prove that the unlabeled nodes that fund subgraphs predicted as suspicious are in fact illicit entities. To that end, investigations were undertaken to identify the origin of funds flowing into subgraphs deemed suspicious, i.e. to identify the entity controlling the node preceding the subgraph. These investigations utilised open source research and other standard identification techniques and resulted in the identification of a number of nodes. For example:
\begin{enumerate}
    \item At least sixty subgraphs deemed suspicious received their funds from a node identified to be a cryptocurrency mixer. Mixers provide obfuscation services and are heavily-used by those laundering proceeds of illicit activity.
    \item Two suspicious subgraphs received their funds from a Panama-based ponzi scheme. 
    \item At least one hundred suspicious subgraphs received their funds from a node identified to be a bot that enables anonymous cryptocurrency trading on a messaging platform. It is functionally similar to a service offered by the cryptocurrency exchange Bitzlato, which was shut down after almost half of its transactions were linked to criminal activities \cite{Europol_2023}.
    \item At least twenty suspicious subgraphs received funds from a node believed to be an invite-only Russian darknet market. 
    
\end{enumerate}

These results suggest an application of the model output beyond identification of suspicious transactions, i.e. the identification of suspicious nodes. The model has been shown here to be beneficial in directing manual investigations towards nodes most likely to be illicit, based on the subgraphs originating from them.

%%%%%%%%%%%%%%%%%%%%%%%%%%%%%%%%%%%%%%%%%%%%%%%%%%%%%%%%%%%%%%%%%%%%%%%%%%%%%%%%%%%%%%%%%%%%%%%%%%%%%%%%%%%%%%%%%%%%%%%%%%%%%%%%%%%%%%%
\subsection{Identification of known cryptocurrency laundering patterns}
Further validation of the model results can be obtained by examining the suspicious subgraphs for known cryptocurrency laundering transaction patterns. 

Many of the suspicious subgraphs were found to contain what are known as “peeling chains”. This refers to a transaction pattern created when a cryptocurrency user sends or “peels” cryptocurrency to a destination address, while the remainder is sent to another address under the user’s control. This happens repeatedly to form a peeling chain. The pattern can have legitimate financial privacy purposes, for example to avoid address reuse, which can enable a user's transactions to be easily linked together. However the pattern can also be indicative of money laundering, as described in various criminal cases \cite{Affidavit_2020}, especially where the “peeled” cryptocurrency is repeatedly sent to an exchange service. In traditional finance this is known as “smurfing”, where large amounts of cash are structured into multiple small transactions, to keep them under regulatory reporting limits and avoid detection. Given the model was using node degree, it is understandable that peeling chains are a pattern visible to it, as a peeling chain is likely to result in a chain of nodes all with matching node degree.

Aside from peeling chains, many of the suspicious subgraphs contain apparent “nested services” in the vicinity of the final deposits into cryptocurrency exchanges. Nested services are businesses that move funds through accounts at larger cryptocurrency exchanges, sometimes without the awareness or approval of the exchange. A nested service might receive a deposit from one of their customers into a cryptocurrency address, and then forward the funds to their deposit address at an exchange. Nested services are known to frequently have less stringent customer due diligence checks than the cryptocurrency exchanges they utilise, or sometimes have no such anti-money laundering checks at all, resulting in their misuse for cryptocurrency laundering \cite{Elliptic_nest} - potentially causing them to feature in subgraphs deemed by the model as suspicious.

%%%%%%%%%%%%%%%%%%%%%%%%%%%%%%%%%%%%%%%%%%%%%%%%%%%%%%%%%%%%%%%%%%%%%%%%%%%%%%%%%%%%%%%%%%%%%%%%%%%%%%%%%%%%%%%%%%%%%%%%%%%%%%%%%%%%%%%
\section{Subgraph learning at scale}\label{sec:scale}
The large size of \ourdataset poses substantial challenges for training GNNs. It is imperative to build scalable training and inference systems that address the bottlenecks and workloads of subgraph representation learning. We use GLASS~\cite{glass} as a motivating example, because it is a simple adaptation of node-classification GNNs while being provably expressive and empirically effective. Specifically, GLASS appends an extra, binary attribute to the feature vector, indicating whether a node belongs to the subgraphs of interest. The authors show that such a \emph{0-1 labeling trick} enables learning any function over subgraphs, provided that the backbone GNN is sufficiently expressive.

Subgraph methods like GLASS require some changes to those for node classification. Here, we introduce a scalable GNN training system, SALIENT~\cite{kaler2022accelerating}, together with its improved successor SALIENT++~\cite{kaler2023communication}, which was designed for node classification workloads. Then, we discuss how systems like SALIENT/SALIENT++ can be adapted to address subgraph classification workloads.

%%%%%%%%%%%%%%%%%%%%%%%%%%%%%%%%%%%%%%%%%%%%%%%%%%%%%%%%%%%%%%%%%%%%%%%%%%%%%%%%%%%%%%%%%%%%%%%%%%%%%%%%%%%%%%%%%%%%%%%%%%%%%%%%%%%%%%%
\subsection{SALIENT and SALIENT++}
Large-scale, distributed training of GNNs is faced with two major bottlenecks: the cost of neighborhood sampling dominates that of model evaluation and the distributed storage of node features incurs heavy inter-machine communications.

Neighborhood sampling is a means to reduce the explosive size of the $k$-hop neighborhood of a minibatch of training nodes. This strategy reduces the computation and memory required in minibatch training. A commonly employed neighborhood sampling algorithm is called \emph{nodewise sampling} wherein the sampled $k$-hop neighborhood is computed by sampling up to $f_k$ neighbors for each node in the $(k-1)$-hop neighborhood. Neighborhood sampling is performed in CPU memory, as a main computational component of the data loader; however, the CPU throughput is way below that of GPUs, which perform model evaluations based on the sampled neighbors.

To address the bottleneck of neighborhood sampling, SALIENT~\cite{kaler2022accelerating} implements the sampler in C++ by using the most efficient data structures, performs shared-memory parallel batch preparation by using C++ threads as opposed to PyTorch's multi-processing, and pipelines the data transfers between GPUs and CPUs to maximize GPU utilization. These performance engineerings enable SALIENT to achieve a speedup of 3$\times$ over the standard GNN system implemented by using the popular PyTorch-Geometric library with a single GPU and a further 8$\times$ parallel speedup with 16 GPUs, on the ogbn-papers100M dataset, one of the largest benchmarks for graph deep learning.

When node features are partitioned (such that each partition contains a subset of nodes) and stored in different machines, the communication of features across machines becomes another training bottleneck. Inter-machine communication occurs because the sampled $k$-hop neighborhood inevitably includes nodes outside a partition (i.e., those stored in another machine).

To reduce the communication volume, SALIENT++~\cite{kaler2023communication} employs a static caching policy that caches the features of the frequently accessed out-of-partition nodes. This policy is based on a technique called \emph{vertex-inclusion probability} (VIP) analysis, which works in the following. Given a graph $G=(V,E)$ and $\mathcal{T} \subset V$ of vertices in the training set, VIP analysis begins by estimating the probability that a given vertex will be present in a randomly selected minibatch of size $B$. The probability that any given vertex appears in such a minibatch is $|\mathcal{T}| / |V|$. An iterative process is then performed to calculate the probability that a vertex will be sampled, using neighborhood sampling, in each layer of the GNN. This caching policy enables SALIENT++ to achieve a 12.7$\times$ speedup over another popular training system, DistDGL, on eight machines for the ogbn-papers100M benchmark.

%%%%%%%%%%%%%%%%%%%%%%%%%%%%%%%%%%%%%%%%%%%%%%%%%%%%%%%%%%%%%%%%%%%%%%%%%%%%%%%%%%%%%%%%%%%%%%%%%%%%%%%%%%%%%%%%%%%%%%%%%%%%%%%%%%%%%%%
\subsection{Adapting node-classification GNNs to subgraph classification workloads}
For subgraph methods such as GLASS, existing efficient training systems for node classification can be modified to handle subgraph classification workloads. We take SALIENT and SALIENT++ for example.

First, the neighborhood sampling code in SALIENT can be adapted to operate on subgraphs with only minor modifications to the composition of minibatches. A minibatch of subgraphs can be represented as a list of nodes combined with metadata indicating the ranges in the list that correspond to distinct subgraphs. A minibatch can be constructed during training by shuffling the subgraphs in the training set and then forming a minibatch of the selected subgraphs' contained nodes. With these adaptations, the existing fast sampling code in SALIENT can be used to address the computational bottlenecks of neighborhood sampling in subgraph classification.

Second, the VIP analysis technique used by SALIENT++ can be adapted to handle subgraph workloads in a fairly straightforward manner as well. Conceptually, one can reduce the problem to that of computing an optimal static caching policy for a node classification workload with an augmented graph and sampling scheme. The augmented graph $G'$ contains all of the nodes and edges of $G$, but additionally contains one extra node per subgraph which is connected to all nodes in $G$ contained in that subgraph by directed edges. The augmented sampling scheme includes one extra layer that samples all neighbors of the ``subgraph nodes.'' VIP analysis may now be performed on $G'$ by treating the set of added ``subgraph nodes'' as the training set of a node classification workload. In practice, there is no need to actually materialize $G'$ as the original graph $G$ combined with a list of subgraphs in $G$ is sufficient to perform the required update.

%%%%%%%%%%%%%%%%%%%%%%%%%%%%%%%%%%%%%%%%%%%%%%%%%%%%%%%%%%%%%%%%%%%%%%%%%%%%%%%%%%%%%%%%%%%%%%%%%%%%%%%%%%%%%%%%%%%%%%%%%%%%%%%%%%%%%%%
\section{Summary}
We have contributed a second large, labeled cryptocurrency transaction data set, \ourdataset, five years after \textit{Elliptic1} was published. 
%
% To anonymize identity
% We have contributed a large, labeled cryptocurrency transaction data set, \ourdataset, five years after \textit{Elliptic1} was published. 
\ourdataset is more than two orders of magnitude larger than \textit{Elliptic1} and it supports a different machine-learning task -- subgraph classification -- for AML. It offers research opportunities in not only financial forensics but also machine learning, particularly, subgraph representation learning. We demonstrated its utility through training subgraph-based predictive models and discovered an effective one: GLASS.

The model produced very promising results, identifying illicit activity based on on-chain patterns, which were previously only identifiable through off-chain information, once funds entered a regulated exchange service. This opens the way to using the output of these models as an effective compliance tool.

The model also successfully directed us towards previously unknown illicit cryptocurrency wallets, based on the subgraphs - i.e. based on the way funds from these wallets were being laundered. This could be utilised by blockchain analytics companies, law enforcement investigators and financial regulators to identify cryptocurrency wallets involved in illicit activity.

While the model applied here has demonstrated promising results, future studies can improve the prediction performance by using a scalable training system that incorporates node/edge features into the model under a computational budget. The inclusion of more features could enable discovery of more sophisticated and previously undocumented money laundering strategies, leading to the more accurate identification of financial crime.

%%%%%%%%%%%%%%%%%%%%%%%%%%%%%%%%%%%%%%%%%%%%%%%%%%%%%%%%%%%%%%%%%%%%%%%%%%%%%%%%%%%%%%%%%%%%%%%%%%%%%%%%%%%%%%%%%%%%%%%%%%%%%%%%%%%%%%%
\section{Data and code}
The \ourdataset dataset can be accessed through \url{http://elliptic.co/elliptic2}. The accompanying code used for experimentation is located at \url{https://github.com/MITIBMxGraph/Elliptic2}.
% The \ourdataset dataset and the accompanying code used for experimentation will be revealed on acceptance of the paper.

%%%%%%%%%%%%%%%%%%%%%%%%%%%%%%%%%%%%%%%%%%%%%%%%%%%%%%%%%%%%%%%%%%%%%%%%%%%%%%%%%%%%%%%%%%%%%%%%%%%%%%%%%%%%%%%%%%%%%%%%%%%%%%%%%%%%%%%
%
%ACKNOWLEDGEMENTS
\begin{acks}
This work was funded by Elliptic and the MIT-IBM Watson AI Lab, a joint research initiative between the Massachusetts Institute of Technology and IBM Research. Data and domain expertise were provided by Elliptic.
\end{acks}

%%%%%%%%%%%%%%%%%%%%%%%%%%%%%%%%%%%%%%%%%%%%%%%%%%%%%%%%%%%%%%%%%%%%%%%%%%%%%%%%%%%%%%%%%%%%%%%%%%%%%%%%%%%%%%%%%%%%%%%%%%%%%%%%%%%%%%%
%
% The next two lines define the bibliography style to be used, and the bibliography file.
\bibliographystyle{ACM-Reference-Format}
\bibliography{bibliography}
% 
% If your work has an appendix, this is the place to put it.
%\appendix

\end{document}